\documentclass[a4paper,11pt]{article}

\usepackage{colacl}
\usepackage{times}

\title{Automatic Extraction of Subcategorization Frames for
Czech\thanks{ This work was done during the second author's visit to
the University of Pennsylvania. We would like to thank Prof. Aravind
Joshi, David Chiang, Mark Dras and the anonymous reviewers for their
comments. The first author's work is partially supported by NSF Grant
SBR 8920230.  Many tools used in this work are the results of project
No. VS96151 of the Ministry of Education of the Czech Republic. The
data (PDT) is thanks to grant No. 405/96/K214 of the Grant Agency of
the Czech Republic. Both grants were given to the Institute of Formal
and Applied Linguistics, Faculty of Mathematics and Physics, Charles
University, Prague. }}

\author{Anoop Sarkar \\[2pt]
Department of Computer and Info. Sci. \\
University of Pennsylvania \\
200 South 33rd Street, \\
Philadelphia, PA 19104 USA \\
{\tt anoop@linc.cis.upenn.edu}
\And
Daniel Zeman\\
\'Ustav form\'aln\'{\i} a aplikovan\'e lingvistiky \\
Univerzita Karlova \\
Malostransk\'e n\'am\v{e}st\'{\i} 25 \\
CZ-11800  Praha, Czechia \\
{\tt zeman@ufal.mff.cuni.cz}}

\usepackage{graphicx}
\usepackage{psfrag}

\newcounter{sentencectr}
\newcounter{sentencesubctr}

\renewcommand{\thesentencectr}{(\smainform{sentencectr})}
\renewcommand{\thesentencesubctr}{\thesentencectr\ssubform{sentencesubctr}}

\newcommand{\smainform}{\arabic}
\newcommand{\ssubform}{\alph}
\newcommand{\ssubpunc}{.{}}

\newcommand{\beginsentences}{ %
\pagebreak[3] %
\begin{list}{(\thesentencectr)}
   {\usecounter{sentencesubctr}
    \setlength{\topsep}{1ex}
    \setlength{\itemsep}{0 in}
    \setlength{\labelwidth}{0.5 in}
    \addtolength{\leftmargin}{4ex}
    \setlength{\labelsep}{.05in}
    \setlength{\parsep}{0 in}}}
\def\endsentences{\end{list}}

\newcommand{\smainitem}{\renewcommand{\thesentencesubctr
                                    }{\thesentencectr\ssubform{sentencesubctr}}
                        \setcounter{sentencesubctr}{0}
                        \refstepcounter{sentencectr}
                        \refstepcounter{sentencesubctr}
     \item[\thesentencectr\hfill\ssubform{sentencesubctr}\ssubpunc]}
\newcommand{\ssubitem}{\refstepcounter{sentencesubctr}
     \item[\hfill\ssubform{sentencesubctr}\ssubpunc]}

\makeatletter
\newcommand{\smainlabel}[1]{{
\renewcommand{\@currentlabel}{\thesentencectr}\label{#1}}}

\newcommand{\ssublabel}[1]{{
\renewcommand{\@currentlabel}{\ssubform{sentencesubctr}}\label{#1}}}
\makeatother

\makeatother

\newcommand{\sep}{\,\mid\,}
\newcommand{\comment}[1]{}

\begin{document}
\maketitle

\begin{abstract}
We present some novel machine learning techniques for the
identification of subcategorization information for verbs in Czech. We
compare three different statistical techniques applied to this
problem.  We show how the learning algorithm can be used to discover
previously unknown subcategorization frames from the Czech Prague
Dependency Treebank. The algorithm can then be used to label
dependents of a verb in the Czech treebank as either arguments or
adjuncts. Using our techniques, we are able to achieve 88\% precision
on unseen parsed text.
\end{abstract}

\section{Introduction}

The subcategorization of verbs is an essential issue in parsing,
because it helps disambiguate the attachment of arguments and recover
the correct predicate-argument relations by a parser.
\cite{carroll98:_subcat_help_parser,carroll98:_valen_pcfg} give several
reasons why subcategorization information is important for a natural
language parser. Machine-readable dictionaries are not comprehensive
enough to provide this lexical
information~\cite{manning93:_subcat,briscoe97:_subcat}. Furthermore,
such dictionaries are available only for very few languages. We need
some general method for the automatic extraction of subcategorization
information from text corpora.

Several techniques and results have been reported on learning
subcategorization frames (SFs) from text corpora
\cite{webster89:_lexical_frames,brent91:_subcat,brent93:_unsup_learn,brent94:_acquis_subcat,ushioda93:_verb_subcat,manning93:_subcat,ersan96:_case_frames,briscoe97:_subcat,carroll98:_subcat_help_parser,carroll98:_valen_pcfg}.
All of this work deals with English. In this paper we report on
techniques that automatically extract SFs for Czech, which is a free
word-order language, where verb complements have visible case
marking.\footnote{ One of the anonymous reviewers pointed out
that~\cite{basili98:_subcat} presents a corpus-driven acquisition of
subcategorization frames for Italian. }

Apart from the choice of target language, this work also differs from previous
work in other ways. Unlike all other previous work in this area, we do
not assume that the set of SFs is known to us in advance. Also in
contrast, we work with syntactically annotated data (the Prague
Dependency Treebank, PDT~\cite{hajic98:_pdt}) where the
subcategorization information is {\em not} given; although this might
be considered a simpler problem as compared to using raw text, we have
discovered interesting problems that a user of a raw or tagged corpus
is unlikely to face.

We first give a detailed description of the task of uncovering SFs and
also point out those properties of Czech that have to be taken into
account when searching for SFs. Then we discuss some differences from
the other research efforts. We then present the three techniques that
we use to learn SFs from the input data.

In the input data, many observed dependents of the verb are
adjuncts. To treat this problem effectively, we describe a novel
addition to the hypothesis testing technique that uses subset of
observed frames to permit the learning algorithm to better distinguish
arguments from adjuncts.

Using our techniques, we are able to achieve 88\% precision in
distinguishing arguments from adjuncts on unseen parsed text.

\section{Task Description}

In this section we describe precisely the proposed task. We also
describe the input training material and the output produced by our
algorithms.

\subsection{Identifying subcategorization frames}
\label{sec:expl}

In general, the problem of identifying subcategorization frames is to
distinguish between arguments and adjuncts among the constituents
modifying a verb. e.g., in ``John saw Mary yesterday at the station'',
only ``John'' and ``Mary'' are required arguments while the other
constituents are optional (adjuncts). There is some controversy as to
the {\em correct} subcategorization of a given verb and linguists
often disagree as to what is the right set of SFs for a given verb. A
machine learning approach such as the one followed in this paper
sidesteps this issue altogether, since it is left to the algorithm to
learn what is an appropriate SF for a verb.

\begin{figure*}[htbp]
  \begin{center}
    \leavevmode
    \psfrag{[#, ZSB, 0]}{\small [\# ZSB 0]}
    \psfrag{[maji, VPP3A, 2]}{\small [maj\'{\i} VPP3A 2]}
    \psfrag{[zajem, N4, 5]}{\small [z\'ajem N4 5]}
    \psfrag{[o, R4, 3]}{\small [o R4 3]}
    \psfrag{[jazyky, NIP4A, 4]}{\small [jazyky N4 4]}
    \psfrag{[fakulte, N3, 7]}{\small [fakult\v{e} N3 7]}
    \psfrag{[vsak, JE, 8]}{\small [v\v{s}ak JE 8]}
    \psfrag{[chybi, VPP3A, 9]}{\small [chyb\'{\i} VPP3A 9]}
    \psfrag{[anglictinari, N1, 10]}{\small [angli\v{c}tin\'a\v{r}i N1 10]}
    \psfrag{[studenti, N1, 1]}{\small [studenti N1 1]}
    \psfrag{[\\,, ZIP, 6]}{\small [, ZIP 6]}
    \psfrag{[., ZIP, 11]}{\small [. ZIP 11]}
    \psfrag{have}{\small have}
    \psfrag{but}{\small but}
    \psfrag{students}{\small students}
    \psfrag{faculty(dative)}{\small faculty(dative)}
    \psfrag{teachers of english}{\small teachers of English}
    \psfrag{miss}{\small miss}
    \psfrag{interest}{\small interest}
    \psfrag{in}{\small in}
    \psfrag{languages}{\small languages}
    \includegraphics[height=3in]{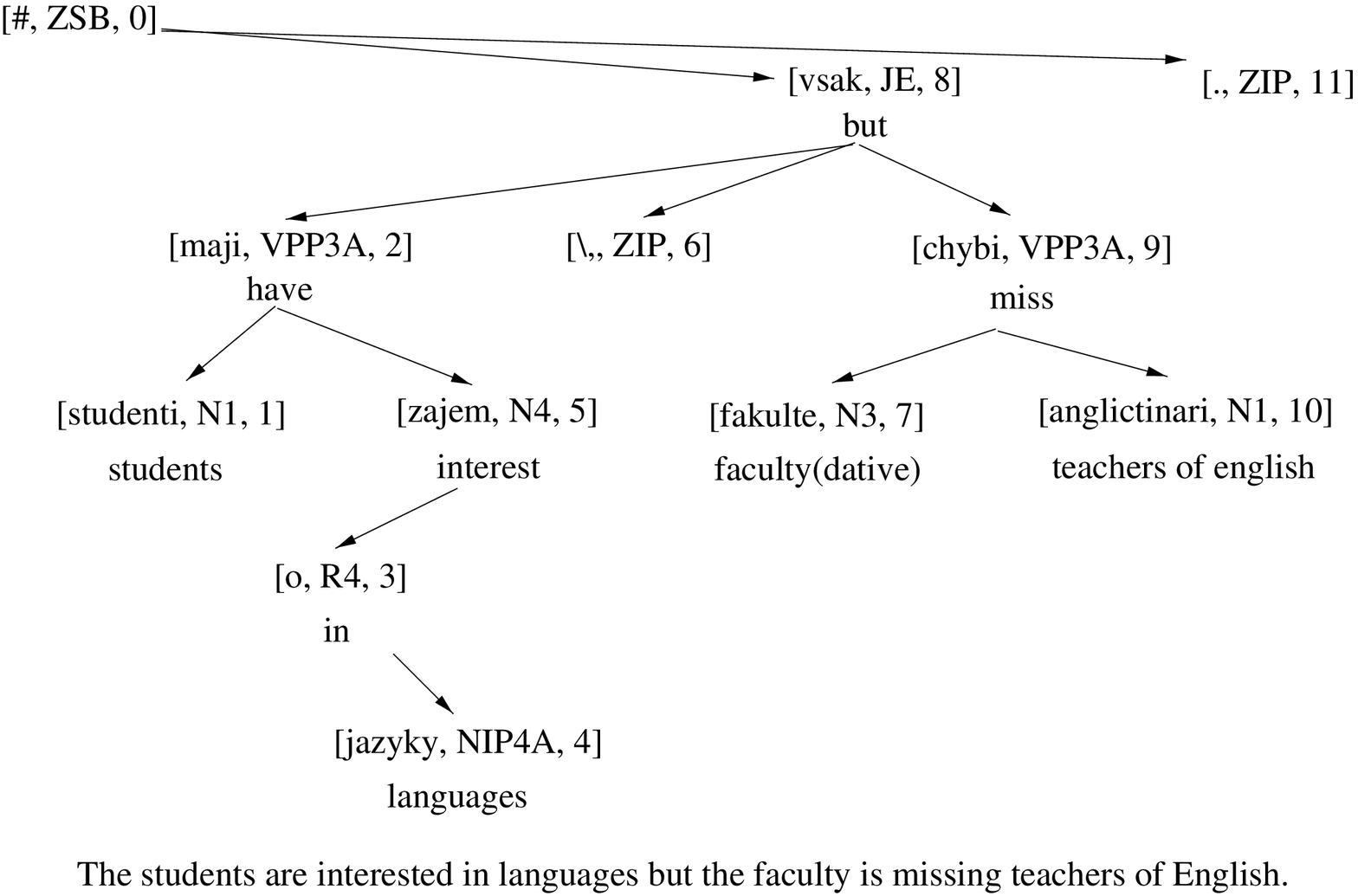}
    \caption{Example input to the algorithm from the Prague Dependency Treebank}
    \label{fig:pdt_ex}
  \end{center}
\end{figure*}

Figure~\ref{fig:pdt_ex} shows a sample input sentence from the PDT
annotated with dependencies which is used as training material for the
techniques described in this paper. Each node in the tree contains a
word, its part-of-speech tag (which includes morphological
information) and its location in the sentence. We also use the
functional tags which are part of the PDT annotation\footnote{ For
those readers familiar with the PDT functional tags, it is important
to note that the functional tag {\em Obj} does not always correspond
to an argument. Similarly, the functional tag {\em Adv} does not
always correspond to an adjunct. Approximately 50 verbs out of the
total 2993 verbs require an adverbial argument.}. To make future
discussion easier we define some terms here. Each daughter of a verb
in the tree shown is called a {\em dependent} and the set of all
dependents for that verb in that tree is called an {\em observed frame
(OF)}. A {\em subcategorization frame (SF)} is a subset of the OF. For
example the OF for the verb {\em maj\'{\i} (have)} in
Figure~\ref{fig:pdt_ex} is {\em \{~N1, N4~\}} and its SF is the same
as its OF. Note that which OF (or which part of it) is a true SF is not
marked in the training data. After training on such examples, the
algorithm takes as input parsed text and labels each daughter of each
verb as either an argument or an adjunct. It does this by selecting the
most likely SF for that verb given its OF.


\subsection{Relevant properties of the Czech Data}

Czech is a ``free word-order'' language. This means that the
arguments of a verb do not have fixed positions and are not guaranteed
to be in a particular configuration with respect to the verb.

The examples in \ref{ex:wordorder} show that while Czech has a
relatively free word-order some orders are still marked. The SVO, OVS,
and SOV orders in \ref{ex:svo}, \ref{ex:ovs}, \ref{ex:sov}
respectively, differ in emphasis but have the same predicate-argument
structure. The examples \ref{ex:ques1}, \ref{ex:ques2} can only be
interpreted as a question. Such word orders require proper intonation
in speech, or a question mark in text.


The example \ref{ex:morphmarking} demonstrates how morphology is
important in identifying the arguments of the
verb. cf. \ref{ex:morphmarking} with \ref{ex:ovs}. The ending {\em -a}
of {\em Martin} is the only difference between the two sentences. It
however changes the morphological case of {\em Martin} and turns it
from subject into object.  Czech has 7 cases that can be distinguished
morphologically.

\beginsentences
\smainitem Martin otv\'{\i}r\'a soubor. (SVO: Martin opens the file) \label{ex:svo}
\ssubitem  Soubor otv\'{\i}r\'a Martin. (OVS: $\neq$ the file opens Martin) \label{ex:ovs}
\ssubitem  Martin soubor otv\'{\i}r\'a. \label{ex:sov}
\ssubitem \#Otv\'{\i}r\'a Martin soubor. \label{ex:ques1}
\ssubitem \#Otv\'{\i}r\'a soubor Martin. \label{ex:ques2}
\ssubitem Soubor otv\'{\i}r\'a Martina. ($=$ the file opens Martin) \label{ex:morphmarking}
\smainlabel{ex:wordorder}
\endsentences

Almost all the existing techniques for extracting SFs exploit the
relatively fixed word-order of English to collect features for their
learning algorithms using fixed patterns or rules (see
Table~\ref{tbl:previous_work} for more details). Such a technique is
not easily transported into a new language like Czech.
Fully parsed training data can help here by supplying all dependents
of a verb.
The observed frames obtained this way have to be {\em normalized} with
respect to the word order, e.g. by using an alphabetic ordering.

For extracting SFs, prepositions in Czech have to be handled
carefully. In some SFs, a particular preposition is required by the
verb, while in other cases it is a class of prepositions such as
locative prepositions (e.g. {\em in, on, behind, $\ldots$}) that are
required by the verb. In contrast, adjuncts can use a wider variety of
prepositions. Prepositions specify the case of their noun phrase
complements but a preposition can take complements with more than one
case marking with a different meaning for each case. (e.g. {\em na
most\v{e} $=$ on the bridge; na most $=$ onto the bridge}). In general,
verbs select not only for particular prepositions but also indicate the
case marking for their noun phrase complements.


\subsection{Argument types}

We use the following set of labels as possible arguments for a verb in
our corpus.  They are derived from morphological tags and simplified
from the original PDT definition~\cite{hajic98:_tagger,hajic98:_pdt};
the numeric attributes are the case marking identifiers.  For
prepositions and clause complementizers, we also save the lemma in
parentheses.

\begin{itemize}

\item Noun phrases: N4, N3, N2, N7, N1

\item Prepositional phrases: R2(bez), R3(k), R4(na), R6(na), R7(s),
$\ldots$

\item Reflexive pronouns {\em se}, {\em si}: PR4, PR3

\item Clauses: S, JS(\v{z}e), JS(zda)

\item Infinitives (VINF)

\item passive participles (VPAS)

\item adverbs (DB)

\end{itemize}

We do not specify which SFs are possible since we aim to discover
these (see Section~\ref{sec:expl}).

\section{Three methods for identifying subcategorization frames}
\label{sec:methods}

We describe three methods that take as input a list of verbs and
associated observed frames from the training data (see
Section~\ref{sec:expl}), and learn an association between verbs and
possible SFs. We describe three methods that arrive at a numerical
score for this association.

However, before we can apply any statistical methods to the training
data, there is one aspect of using a treebank as input that has to be
dealt with. A correct frame (verb + its arguments) is almost always
accompanied by one or more adjuncts in a real sentence. Thus the {\em
observed frame} will almost always contain noise. The approach offered
by Brent and others counts all observed frames and then decides which
of them do not associate strongly with a given verb. In our situation
this approach will fail for most of the observed frames because we
rarely see the correct frames isolated in the training data. For example,
from the occurrences of the transitive verb {\it absolvovat} (``go through
something'') that occurred ten times in the corpus, no occurrence
consisted of the verb-object pair alone. In other words, the correct SF
constituted 0\% of the observed situations. Nevertheless, for each
observed frame, one of its subsets was the correct frame we sought
for. Therefore, we considered all possible subsets of all observed
frames. We used a technique which steps through the subsets of each
observed frame from larger to smaller ones and records their frequency
in data.  Large infrequent subsets are suspected to contain adjuncts,
so we replace them by more frequent smaller subsets. Small infrequent
subsets may have elided some arguments and are rejected. Further
details of this process are discussed in Section~\ref{sec:miscue}. 

\begin{figure*}[htbp]
  \begin{center}
    \leavevmode
    \includegraphics[height=1.6in]{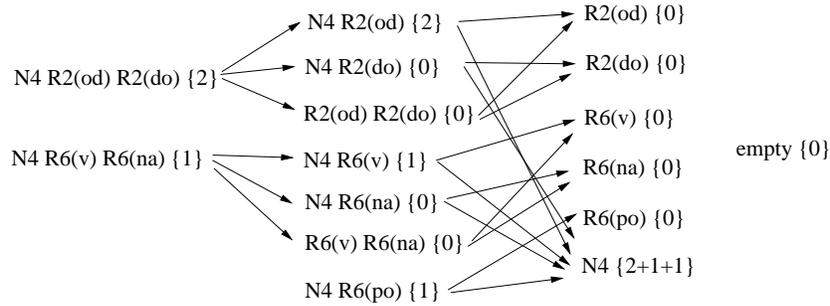}

    \caption{Computing the subsets of observed frames for the verb
    {\em \ absolvovat}. The counts for each frame are given within
    braces $\{\}$. In this example, the frames {\em N4 R2(od), N4
    R6(v)} and {\em N4 R6(po)} have been observed with other verbs in
    the corpus. Note that the counts in this figure do not correspond
    to the real counts for the verb {\em absolvovat} in the training
    corpus.}

    \label{fig:subsets}
  \end{center}
\end{figure*}

The methods we present here have a common structure. For each verb, we
need to associate a score to the hypothesis that a particular set of
dependents of the verb are arguments of that verb. In other words, we
need to assign a value to the hypothesis that the observed frame under
consideration is the verb's SF. Intuitively, we either want to test
for independence of the observed frame and verb distributions in the
data, or we want to test how likely is a frame to be observed with a
particular verb without being a valid SF. We develop these intuitions
with the following well-known statistical methods. For further
background on these methods the reader is referred to
\cite{bickel77:_mathem_statis,dunning93:_statis}.

\subsection{Likelihood ratio test}
\label{sec:lik}

Let us take the hypothesis that the distribution of an observed frame
$f$ in the training data is independent of the distribution of a verb
$v$. We can phrase this hypothesis as $p(f \sep v) = p(f \sep\ !v) =
p(f)$, that is distribution of a frame $f$ given that a verb $v$ is
present is the same as the distribution of $f$ given that $v$ is not
present (written as $!v$). We use the log likelihood test
statistic~\cite{bickel77:_mathem_statis}(p.209) as a measure to
discover particular frames and verbs that are highly associated in the
training data. 

\begin{eqnarray}
  \label{eqn:lik1}
	k_1 & = & c(f,v) \nonumber \\
	n_1 & = & c(v) = c(f,v) + c(!f,v) \nonumber \\
	k_2 & = & c(f,!v) \nonumber \\
	n_2 & = & c(!v) = c(f,!v) + c(!f,!v) \nonumber
\end{eqnarray}

where $c(\cdot)$ are counts in the training data. Using the values
computed above:

\begin{eqnarray}
  \label{eqn:lik2}
	p_1 & = & \frac{k_1}{n_1}  \nonumber \\
	p_2 & = & \frac{k_2}{n_2} \nonumber \\
	p & = & \frac{k_1 + k_2}{n_1 + n_2} \nonumber
\end{eqnarray}

Taking these probabilities to be binomially distributed, the log
likelihood statistic~\cite{dunning93:_statis} is given by:

\begin{eqnarray}
  \label{eqn:lik3}
\lefteqn{-2 \log \lambda = } \nonumber \\
&& 2 [ \log L(p_1, k_1, n_1) + \log L(p_2, k_2, n_2) - \nonumber \\
&&\ \ \ \log L(p, k_1, n_2) - \log L(p, k_2, n_2) ] \nonumber 
\end{eqnarray}

where,

\[ \log L(p,n,k) = k \log p + (n - k) \log(1 - p) \]

According to this statistic, the greater the value of $-2 \log
\lambda$ for a particular pair of observed frame and verb, the more
likely that frame is to be valid SF of the verb.

\subsection{T-scores}

Another statistic that has been used for hypothesis testing is the
{\em t-score}. Using the definitions from Section~\ref{sec:lik} we can
compute t-scores using the equation below and use its value to measure
the association between a verb and a frame observed with it. 

\[ T = \frac{ p_1 - p_2 }
	{ \sqrt{ \sigma^2(n_1, p_1) + \sigma^2(n_2, p_2) } } \]

where,

\[ \sigma(n,p) = n p ( 1 - p ) \]

In particular, the hypothesis being tested using the t-score is
whether the distributions $p_1$ and $p_2$ are {\em not}
independent. If the value of $T$ is greater than some threshold then
the verb $v$ should take the frame $f$ as a SF.

\subsection{Binomial Models of Miscue Probabilities}
\label{sec:miscue}

Once again assuming that the data is binomially distributed, we can
look for frames that co-occur with a verb by exploiting the miscue
probability: the probability of a frame co-occuring with a verb when
it is not a valid SF. This is the method used by several earlier
papers on SF extraction starting
with~\cite{brent91:_subcat,brent93:_unsup_learn,brent94:_acquis_subcat}.

Let us consider probability $p_{!f}$ which is the probability that a
given verb is observed with a frame but this frame is not a valid SF
for this verb. $p_{!f}$ is the error probability on identifying a SF
for a verb. Let us consider a verb $v$ which does {\em not} have as
one of its valid SFs the frame $f$. How likely is it that $v$ will be
seen $m$ or more times in the training data with frame $f$? If $v$ has
been seen a total of $n$ times in the data, then $H^\ast(p_{!f}; m,
n)$ gives us this likelihood.

\[ H^\ast(p_{!f}; m, n) = 
	\sum_{i = m}^{n} p_{!f}^i ( 1 - p_{!f})^{n - i} 
	\left( \begin{array}{c} n \\ i \end{array} \right) 
\]

If $H^\ast(p; m, n)$ is less than or equal to some small threshold
value then it is extremely unlikely that the hypothesis is true, and
hence the frame $f$ must be a SF of the verb $v$. Setting the
threshold value to $0.05$ gives us a 95\% or better confidence value
that the verb $v$ has been observed often enough with a frame $f$ for
it to be a valid SF.

Initially, we consider only the observed frames (OFs) from the
treebank. There is a chance that some are subsets of some others but
now we count only the cases when the OFs were seen themselves. Let's
assume the test statistic rejected the frame. Then it is not a real SF
but there probably is a subset of it that is a real SF. So we select
exactly one of the subsets whose length is one member less: this is
the {\em successor} of the rejected frame and inherits its
frequency. Of course one frame may be successor of several longer
frames and it can have its own count as OF. This is how frequencies
accumulate and frames become more likely to survive.  The example
shown in Figure~\ref{fig:subsets} illustrates how the subsets and
successors are selected.

An important point is the selection of the successor. We have to
select only one of the $n$ possible successors of a frame of length
$n$, otherwise we would break the total frequency of the verb. Suppose
there is $m$ rejected frames of length $n$. This yields $m * n$
possible modifications to consider before selection of the
successor. We implemented two methods for choosing a single successor
frame:

\begin{enumerate}

\item Choose the one that results in the strongest preference for some
frame (that is, the successor frame results in the lowest entropy
across the corpus). This measure is sensitive to the frequency of this
frame in the rest of corpus.

\item Random selection of the successor frame from the alternatives.

\end{enumerate}

Random selection resulted in better precision (88\% instead of
86\%). It is not clear why a method that is sensitive to the frequency
of each proposed successor frame does not perform better than random
selection.

The technique described here may sometimes result in subset of a
correct SF, discarding one or more of its members. Such frame can
still help parsers because they can at least look for the dependents
that have survived.

\section{Evaluation}

For the evaluation of the methods described above we used the Prague
Dependency Treebank (PDT). We used 19,126 sentences of training data
from the PDT (about 300K words). In this training set, there were
33,641 verb tokens with 2,993 verb types. There were a total of 28,765
{\em observed frames} (see Section~\ref{sec:expl} for explanation of
these terms).
There were 914 verb types seen 5 or more times.

Since there is no electronic valence dictionary for Czech, we
evaluated our filtering technique on a set of 500 test sentences which
were unseen and separate from the training data. These test sentences
were used as a gold standard by distinguishing the arguments and
adjuncts manually. We then compared the accuracy of our output set of
items marked as either arguments or adjuncts against this gold
standard.

First we describe the baseline methods. Baseline method 1: consider
each dependent of a verb an adjunct.  Baseline method 2: use just the
longest known observed frame matching the test pattern. If no matching
OF is known, find the longest partial match in the OFs seen in the
training data. We exploit the functional and morphological tags while
matching. No statistical filtering is applied in either baseline
method.


A comparison between all three methods that were proposed in this
paper is shown in Table~\ref{tbl:comparison}.


\begin{table*}[htbp]
\begin{center}
\begin{tabular}{|l||c|c|c|c|c|} \hline 
  & Baseline 1 & Baseline 2 & Lik. Ratio & T-scores & Hyp. Testing \\
\hline
Precision & 55\% & 78\% & 82\% & 82\% & 88\% \\
Recall: & 55\% & 73\% & 77\% & 77\% & 74\% \\
$F_{\beta=1}$ & 55\% & 75\% & 79\% & 79\% & 80\% \\
\% unknown & 0\% &  6\% &  6\% &  6\% & 16\% \\
\hline
Total verb nodes  & 1027 & 1027 & 1027 & 1027 & 1027 \\
Total complements & 2144 & 2144 & 2144 & 2144 & 2144 \\
Nodes with known verbs & 1027 & 981 & 981 & 981 & 907 \\
Complements of known verbs & 2144 & 2010 & 2010 & 2010 & 1812 \\
Correct Suggestions & 1187.5 & 1573.5 & 1642.5 & 1652.9 & 1596.5 \\
True Arguments & 956.5 & 910.5 & 910.5 & 910.5 & 834.5 \\
Suggested Arguments & 0 & 1122 & 974 & 1026 & 674 \\
Incorrect arg suggestions & 0 & 324 & 215.5 & 236.3 & 27.5 \\
Incorrect adj suggestions & 956.5 & 112.5 & 152 & 120.8 & 188 \\
\hline
\end{tabular}
\caption{Comparison between the baseline methods and the three methods
proposed in this paper. Some of the values are not integers since for
some difficult cases in the test data, the value for each
argument/adjunct decision was set to a value between $[0,1]$. {\em
Recall} is computed as the number of known verb complements divided by
the total number of complements. {\em Precision} is computed as the
number of correct suggestions divided by the number of known verb
complements. $F_{\beta=1} = (2 \times p \times r)/(p+r)$. {\em \%
unknown} represents the percent of test data not considered by a
particular method. }
\label{tbl:comparison}
\end{center}
\end{table*}

The experiments showed that the method improved precision of this
distinction from 57\% to 88\%. We were able to classify as many as 914
verbs which is a number outperformed only by Manning, with 10x more
data (note that our results are for a different language).

Also, our method discovered 137 subcategorization frames from the
data. The known upper bound of frames that the algorithm could have
found (the total number of the {\em observed frame} types) was 450.


\section{Comparison with related work}
\label{sec:relwork}

Preliminary work on SF extraction from corpora was done by
\cite{brent91:_subcat,brent93:_unsup_learn,brent94:_acquis_subcat} and
\cite{webster89:_lexical_frames,ushioda93:_verb_subcat}.
Brent~\cite{brent93:_unsup_learn,brent94:_acquis_subcat} uses the
standard method of testing miscue probabilities for filtering frames
observed with a verb. \cite{brent94:_acquis_subcat} presents a method
for estimating $p_{!f}$. Brent applied his method to a small number of
verbs and associated SF types.  \cite{manning93:_subcat} applies
Brent's method to parsed data and obtains a subcategorization
dictionary for a larger set of
verbs. \cite{briscoe97:_subcat,carroll98:_subcat_help_parser} differs
from earlier work in that a substantially larger set of SF types are
considered; \cite{carroll98:_valen_pcfg} use an EM algorithm to learn
subcategorization as a result of learning rule probabilities, and, in
turn, to improve parsing accuracy by applying the verb SFs obtained.
\cite{basili98:_subcat} use a conceptual clustering algorithm for
acquiring subcategorization frames for Italian. They establish a
partial order on partially overlapping OFs (similar to our OF subsets)
which is then used to suggest a potential SF. A complete comparison of
all the previous approaches with the current work is given in
Table~\ref{tbl:previous_work}.

While these approaches differ in size and quality of training data,
number of SF types (e.g. intransitive verbs, transitive verbs) and
number of verbs processed, there are properties that all have in
common.  They all assume that they know the set of possible SF types
in advance. Their task can be viewed as assigning one or more of the
(known) SF types to a given verb. In addition, except for
\cite{briscoe97:_subcat,carroll98:_subcat_help_parser}, only a small
number of SF types is considered.

Using a dependency treebank as input to our learning algorithm has
both advantages and drawbacks. There are two main advantages of
using a treebank:

\begin{itemize}

\item Access to more accurate data. Data is less noisy when compared
with tagged or parsed input data. We can expect correct identification
of verbs and their dependents.

\item We can explore techniques (as we have done in this paper) that
try and learn the set of SFs from the data itself, unlike other
approaches where the set of SFs have to be set in advance.

\end{itemize}

Also, by using a treebank we can use verbs in different contexts which
are problematic for previous approaches, e.g. we can use verbs that
appear in relative clauses. However, there are two main drawbacks:

\begin{itemize}

\item Treebanks are expensive to build and so the techniques presented
here have to work with less data.

\item All the dependents of each verb are visible to the learning
algorithm. This is contrasted with previous techniques that rely on
finite-state extraction rules which ignore many dependents of the
verb. Thus our technique has to deal with a different kind of data as
compared to previous approaches.

\end{itemize}

We tackle the second problem by using the method of observed frame
subsets described in Section~\ref{sec:miscue}.

\begin{table*}[htbp]
\begin{center}
\begin{tabular}{|l||l|c|c|l|l|l|} \hline
 Previous                    & Data     & \#SFs & \#verbs & Method     & Miscue     & Corpus      \\ 
 work                        &          &       & tested  &            & rate       &             \\ 
\hline
\cite{ushioda93:_verb_subcat}& POS +    & 6     & 33      & heuristics & NA         & WSJ (300K)   \\
                             & FS rules &       &         &            &            &              \\
\hline
\cite{brent93:_unsup_learn}  & raw +    & 6     & 193     & Hypothesis & iterative  & Brown (1.1M) \\ 
                             & FS rules &       &         & testing    & estimation &              \\
\hline
\cite{manning93:_subcat}     & POS +    & 19    & 3104    & Hypothesis & hand       & NYT (4.1M)   \\
                             & FS rules &       &         & testing    &            &              \\
\hline
\cite{brent94:_acquis_subcat}& raw +    & 12    & 126     & Hypothesis & non-iter   & CHILDES (32K)\\
                             & heuristics &     &         & testing    & estimation &              \\
\hline
\cite{ersan96:_case_frames}  & Full     & 16    & 30      & Hypothesis & hand       & WSJ (36M)    \\
                             & parsing  &       &         & testing    &            &              \\
\hline
\cite{briscoe97:_subcat}     & Full     & 160   & 14      & Hypothesis & Dictionary & various (70K)\\
                             & parsing  &       &         & testing    & estimation &              \\
\hline
\cite{carroll98:_valen_pcfg} & Unlabeled & 9+   & 3       & Inside-    & NA         & BNC (5-30M)  \\
                             &          &       &         & outside    &            &              \\
\hline
Current Work                 & Fully    & Learned & 914   & Subsets+   & Estimate   & PDT (300K)   \\
                             & Parsed   & 137     &       & Hyp. testing &          &              \\
\hline
\end{tabular}
\caption{Comparison with previous work on automatic SF extraction from corpora}
\label{tbl:previous_work}
\end{center}
\end{table*}

\section{Conclusion}

We are currently incorporating the SF information produced by the
methods described in this paper into a parser for Czech. We hope to
duplicate the increase in performance shown by treebank-based parsers
for English when they use SF information. Our methods can also be
applied to improve the annotations in the original treebank that we
use as training data. The automatic addition of subcategorization to
the treebank can be exploited to add predicate-argument information to
the treebank.

Also, techniques for extracting SF information from data can be used
along with other research which aims to discover relationships between
different SFs of a
verb~\cite{stevenson99:_verb_class,lapata99:_verb_class,lapata99:_acquir_lexic_gener,stevenson99:_lexical_sem}.

The statistical models in this paper were based on the assumption that
given a verb, different SFs occur independently. This assumption is
used to justify the use of the binomial. Future work perhaps should
look towards removing this assumption by modeling the dependence
between different SFs for the same verb using a multinomial
distribution.

To summarize: we have presented techniques that can be used to learn
subcategorization information for verbs. We exploit a dependency
treebank to learn this information, and moreover we discover the final
set of valid subcategorization frames from the training data. We
achieve upto 88\% precision on unseen data.

We have also tried our methods on data which was automatically
morphologically tagged which allowed us to use more data (82K
sentences instead of 19K). The performance went up to 89\% (a 1\%
improvement).

\bibliographystyle{acl}
{\footnotesize \bibliography{subcat}}
\end{document}